\documentclass[runningheads]{llncs}

\usepackage{eccv}

\usepackage{eccvabbrv}

\usepackage{graphicx}
\usepackage{booktabs}

\usepackage{multirow}
\usepackage{amsmath}
\usepackage{amssymb}
\usepackage{amsfonts}
\usepackage{colortbl}

\usepackage[accsupp]{axessibility}  

\usepackage{hyperref}

\usepackage{orcidlink}

\begin{document}

\title{MV2GF: Multi-view Pedestrian Detection \\ with a Visual Geometric Foundation Model} 

\titlerunning{MV2GF: Multi-view Pedestrian Detection}

\author{
Taiga Yamane\orcidlink{0009-0004-6254-8810} \and
Satoshi Suzuki\orcidlink{0000-0002-1423-3767} \and
Ryo Masumura\orcidlink{0000-0002-2415-4149} \and
Shota Orihashi\orcidlink{0009-0005-5998-6278} \and
Tomohiro Tanaka\orcidlink{0000-0001-8829-2789} \and
Mana Ihori\orcidlink{0009-0003-9114-1174} \and
Naoki Makishima\orcidlink{0000-0002-7065-315X}
}

\authorrunning{T.~Yamane et al.}

\institute{Human Informatics Laboratories, NTT, Inc., Japan \\
\email{\{taiga.yamane, satoshixv.suzuki\}@ntt.com} \\
\url{https://mv2gf.github.io}
}

\maketitle

\begin{abstract}
  Multi-View Pedestrian Detection (MVPD) aims to detect pedestrians in the form of a bird's eye view map from multi-view images.
  Recent MVPD methods adopt a unified framework that projects 2D image features into a 3D world space and aggregates them into a single feature.
  Although they are effective, they struggle to generalize to unseen camera configurations during training due to two main issues.
  First, they are difficult to capture accurate visual geometry across views in unseen camera configurations.
  Second, they make detection models highly dependent on distortion patterns during training arising from their image feature projection.
  To address these, we leverage a visual geometric foundation model and propose MV2GF.
  This foundation model has exhibited strong generalization in capturing visual geometry across views and predicting accurate 3D attributes in diverse camera configurations.
  MV2GF fuses task-specific features with general-purpose geometric features extracted by the foundation model to effectively capture the visual geometry even in unseen camera configurations.
  Furthermore, MV2GF projects each pixel in the image features to an appropriate 3D location using 3D pointmaps predicted by the foundation model, preventing the detection model from depending on distortion patterns during training.
  Our experiments demonstrate the effectiveness of leveraging a visual geometric foundation model for MVPD and that MV2GF generalizes better than existing methods.

  
\end{abstract}

\section{Introduction}
\label{sec:intro}

Pedestrian detection aims to find and localize pedestrians from images.
This task is a crucial component of many applications, such as surveillance systems~\cite{elhoseny2020multi}, autonomous driving~\cite{khan2023localized}, and robotics~\cite{ribeiro2017real}.
These applications often require detection in highly crowded and cluttered scenes.
In such scenes, addressing occlusion, which hinders detection, is important~\cite{chavdarova2018wildtrack}.
Fortunately, multiple cameras with overlapping fields of view are often available in many applications.
Therefore, extensive studies have explored multi-view pedestrian detection (MVPD) to overcome occlusion~\cite{chavdarova2017deep,hou2020multiview,vora2023bringing}.
This task aims to detect pedestrians in the form of a bird's eye view (BEV) map from multi-view images.
By aggregating complementary information across multiple views, MVPD is expected to be more robust to occlusion than single-view pedestrian detection~\cite{dollar2011pedestrian,zhang2017citypersons,braun2019eurocity}.

Recent MVPD methods adopt a unified deep learning-based framework~\cite{hou2020multiview,hou2021multiview,song2021stacked,engilberge2023two,qiu20223d,vora2023bringing,hwang2024booster,aung2024enhancing,suzuki2024scene,zhang2024mahala,aung2024mvpocc,yamane2025msmvd,alturki2025enhanced}.
They first extract 2D image features for individual views and project them into a 3D world space using calibrated camera parameters.
Then, they aggregate projected features from multiple views into a single BEV feature based on visual geometric information across views contained in the image features, such as 3D positional relationships among pedestrians or objects in views.
Finally, they predict a BEV map from the BEV feature.

\begin{figure}[tb]
    \centering
    \includegraphics[width=120mm]{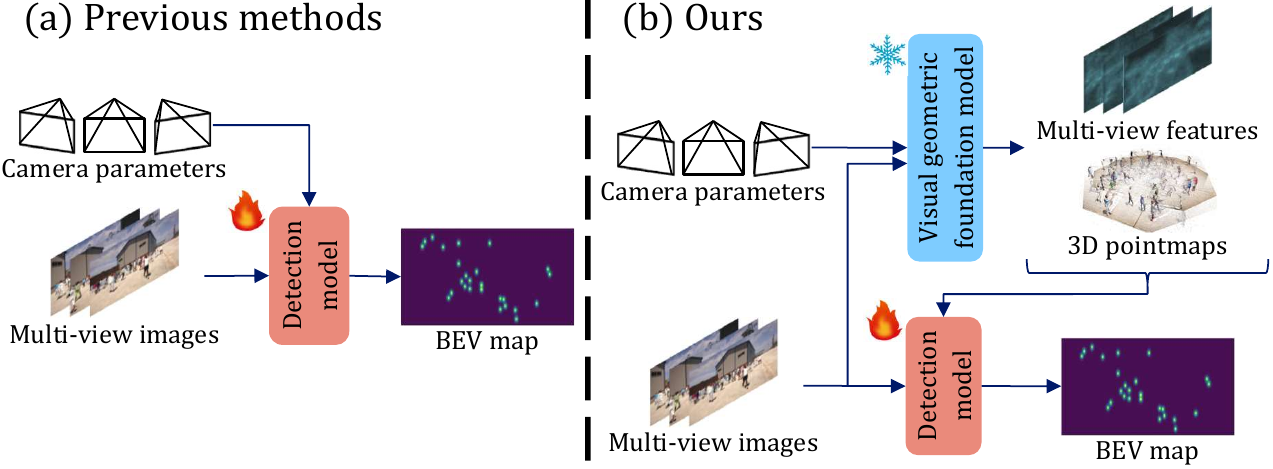}
    \caption{
    (a) Previous MVPD methods. (b) Our MV2GF leverages a pretrained visual geometric foundation model. The red blocks are trainable, and the blue block is frozen.
    }
    \vspace{-4.0mm}
    \label{fig:1}
\end{figure}

While these methods have exhibited promising results, they struggle to generalize to camera configurations not included in MVPD training data~\cite{vora2023bringing,aung2024enhancing}, due to two main issues.
Here, camera configurations include camera types, poses, and positions.
The first issue is that existing methods are difficult to accurately capture visual geometry across views for unseen camera configurations during training.
To accurately capture it, understanding the correspondence between 2D views and the 3D world space is important.
Nonetheless, existing methods learn this correspondence solely from camera configurations in limited MVPD training data.
Since the 2D-3D correspondence varies with camera configurations, for camera configurations not seen during training, existing methods often fail to understand this correspondence and to capture accurate visual geometry.
This makes it difficult for detection models to aggregate features from multiple views for such camera configurations.
The second issue is that existing methods make detection models heavily dependent on distortion patterns arising from image feature projection for camera configurations in MVPD training data.
Existing methods employ a perspective transformation~\cite{hou2020multiview} or its derivative transformations~\cite{song2021stacked,qiu20223d,hwang2024booster,aung2024enhancing,aung2024mvpocc,alturki2025enhanced,yamane2025msmvd} to project 2D image features into a 3D world space.
These transformations cause shadow-like distortions that spread pedestrians out like shadows~\cite{hou2021multiview}, as they project all pixels in image features onto the same planes in a 3D world space.
These distortions obscure precise pedestrian locations on a BEV map, affecting detection performance~\cite{hou2021multiview}.
Existing methods learn distortion patterns from camera configurations in MVPD training data and implicitly compensate for the distortions.
However, they often fail to compensate for them in the patterns for camera configurations not included in the training data, as distortion patterns vary significantly with camera configurations.
This hinders generalization to camera configurations not seen during training.

To overcome these issues, as shown in Fig.~\ref{fig:1}~(b), we leverage a pretrained visual geometric foundation model, also known as a feed-forward 3D reconstruction transformer~\cite{wang2024dust3r,leroy2024grounding,cabon2025must3r,yang2025fast3r,wang2025continuous,tang2025mv,zhang2025flare,wang2025vggt,wang2025pi,depthanything3,keetha2025mapanything}.
It learns to capture visual geometry across views from large-scale multi-view datasets using a transformer~\cite{vaswani2017attention} and demonstrates strong generalization performance in various multi-view 3D tasks across diverse camera configurations.
Therefore, leveraging multi-view image features extracted by the transformer of the foundation model, which contain general-purpose geometric information across views, helps the detection model capture accurate visual geometry for camera configurations not included in MVPD training data.
In addition, instead of using a perspective transformation or its derivatives, we use 3D pointmaps predicted by the foundation model to project image features into a 3D world space.
These pointmaps indicate where each pixel in multi-view images is located in a 3D world space~\cite{wang2024dust3r}.
Using these pointmaps to project image features avoids shadow-like distortions because not all pixels in the image features are projected onto the same planes.

Based on this, we propose \textbf{M}ulti-\textbf{V}iew detection with a \textbf{V}isual \textbf{G}eometric \textbf{F}oundation model, named \textbf{MV2GF}.
In this paper, we mainly employ pretrained Depth Anything 3 (DA3)~\cite{depthanything3} as a visual geometric foundation model.
This is because it can predict 3D attributes at a real-world metric scale, inject camera parameters, and process more than two views in a single forward pass, all of which are suitable for MVPD.
To effectively leverage multi-view image features and 3D pointmaps from DA3, we introduce two key components: Task-specific and Geometric information Fusion (TGF) and Feature Pointmap Aggregation (FPA).
TGF aims to fuse general-purpose geometric information from DA3 with task-specific information and to generate informative multi-view image features.
First, this extracts multi-scale image features for individual views using a trainable ResNet~\cite{he2016deep} and takes multi-view image features from the frozen DA3 transformer.
By training ResNet end-to-end for MVPD, its features contain task-specific information focused on pedestrians, whereas DA3 features contain general-purpose geometric information.
Subsequently, TGF fuses ResNet features with DA3 features using a feature pyramid network (FPN)~\cite{lin2017feature}, yielding informative high-resolution image features.
TGF helps capture visual geometry across views for diverse camera configurations, even those not seen in MVPD training data, while highlighting foreground pedestrians in individual views.
FPA aims to aggregate information across views and generate a BEV feature using image features from TGF and 3D pointmaps from DA3, while avoiding distortions.
First, it projects each pixel in the image features to an appropriate location in a 3D world space using coordinates obtained from the 3D pointmaps.
Then, FPA generates a 3D voxel feature by aggregating features from multiple views within each voxel using max pooling and outputs a BEV feature by compressing the vertical dimension.
FPA prevents the detection model from becoming dependent on distortion patterns during training.

Extensive experiments demonstrate that leveraging a visual geometric foundation model via our TGF and FPA greatly improves generalization performance to camera configurations not included in MVPD training data.
Particularly on MODA metric~\cite{kasturi2008framework}, MV2GF trained on GMVD~\cite{vora2023bringing} training data outperforms the previous state-of-the-art method by $4.6$ points for GMVD testing data, $4.7$ points for MVPerception~\cite{yang2024end}, and $2.2$ points for Wildtrack~\cite{chavdarova2018wildtrack}.

\section{Related Work}

\subsection{Multi-view Pedestrian Detection}
\label{sub:mvpd}

There are two main paradigms in MVPD.
Traditional methods~\cite{baque2017deep,chavdarova2017deep,fleuret2007multicamera,roig2011conditional,xu2016multi,daryani2025camuvid} first detect pedestrians for each view using single-view pedestrian detection~\cite{ren2015faster,zhou2019objects}.
They then aggregate detection results for the same pedestrians across views into locations on a BEV map using techniques such as conditional random field~\cite{baque2017deep}, mean-field inference~\cite{fleuret2007multicamera}, or clustering~\cite{xu2016multi}.
Although these methods are important early attempts, their final detection performance is limited because occlusion severely hinders the detection in each view.
To address this, recent methods~\cite{hou2020multiview,hou2021multiview,song2021stacked,engilberge2023two,qiu20223d,vora2023bringing,hwang2024booster,aung2024enhancing,suzuki2024scene,zhang2024mahala,aung2024mvpocc,yamane2025msmvd,alturki2025enhanced} adopt a unified deep learning-based framework that does not rely on single-view pedestrian detection.
They project image features across multiple views into a 3D world space and aggregate projected features into a single BEV feature based on visual geometric information across views contained in image features.
They predict a BEV map from this BEV feature.
MVDet~\cite{hou2020multiview} is a pioneer of these methods and projects image features onto a ground plane using a perspective transformation.
To further improve detection performance, subsequent studies have extended MVDet by introducing more effective projections~\cite{song2021stacked,qiu20223d,aung2024enhancing,yamane2025msmvd,alturki2025enhanced}, training strategies~\cite{hou2021multiview,engilberge2023two,qiu20223d,vora2023bringing,suzuki2024scene,zhang2024mahala}, and network architectures~\cite{hou2021multiview,lee2023multi,aung2024enhancing,aung2024mvpocc,hwang2024booster}.

Despite these improvements, in terms of capturing visual geometry across views and representing it in image features, many existing methods~\cite{hou2020multiview,song2021stacked,hou2021multiview,qiu20223d,vora2023bringing,aung2024mvpocc,yamane2025msmvd} simply apply a single-view image encoder for each view, such as dilated ResNet~\cite{he2016deep,Yu2016,Yu2017} or FPN~\cite{lin2017feature}.
Some methods~\cite{lee2023multi,hwang2024booster,aung2024enhancing} have introduced effective image feature extraction tailored for MVPD.
MVFP~\cite{aung2024enhancing} highlights the features of the image foreground by combining mean and max pooling, focusing on the geometry of pedestrians.
BoosterSHOT~\cite{hwang2024booster} selects information in image features important for capturing the geometry in scenes using a network like squeeze-and-excitation attention~\cite{hu2018squeeze}.
While they have improved detection performance, existing methods learn to capture visual geometry across views solely from camera configurations in limited MVPD training data and struggle to accurately capture it for camera configurations not seen during training.
Our MV2GF leverages multi-view features extracted by a visual geometric foundation model to effectively capture the visual geometry for diverse camera configurations.

In terms of projecting image features into a 3D world space, existing MVPD methods rely on a perspective transformation~\cite{hou2020multiview} or its derivative transformations~\cite{song2021stacked,qiu20223d,hwang2024booster,aung2024enhancing,aung2024mvpocc,alturki2025enhanced,yamane2025msmvd}.
As described in Sec.~\ref{sec:intro}, these transformations introduce shadow-like distortions~\cite{hou2021multiview}.
By applying a convolutional neural network (CNN) or deformable attention~\cite{zhu2020deformable} to projected features, existing methods learn distortion patterns and implicitly compensate for the distortions, achieving robust detection against patterns identical to those in MVPD training data.
However, this makes detection models heavily dependent on distortion patterns during training and hinders generalization to camera configurations not included in the training data, as distortion patterns vary significantly with camera configurations. 
In the context of autonomous driving, transformer-based projections~\cite{li2022bevformer,huang2023tri,wei2023surroundocc,liu2023fully,li2024viewformer} are the dominant projections and can mitigate the distortions.
However, they assume that camera configurations and the size of the BEV map are identical during training and testing, making them unsuitable for practical MVPD.
Our MV2GF avoids the distortions by using 3D pointmaps predicted by a visual geometric foundation model to project image features, as described in Sec.~\ref{sec:intro}.

\begin{figure}[tb]
    \centering
    \includegraphics[width=120mm]{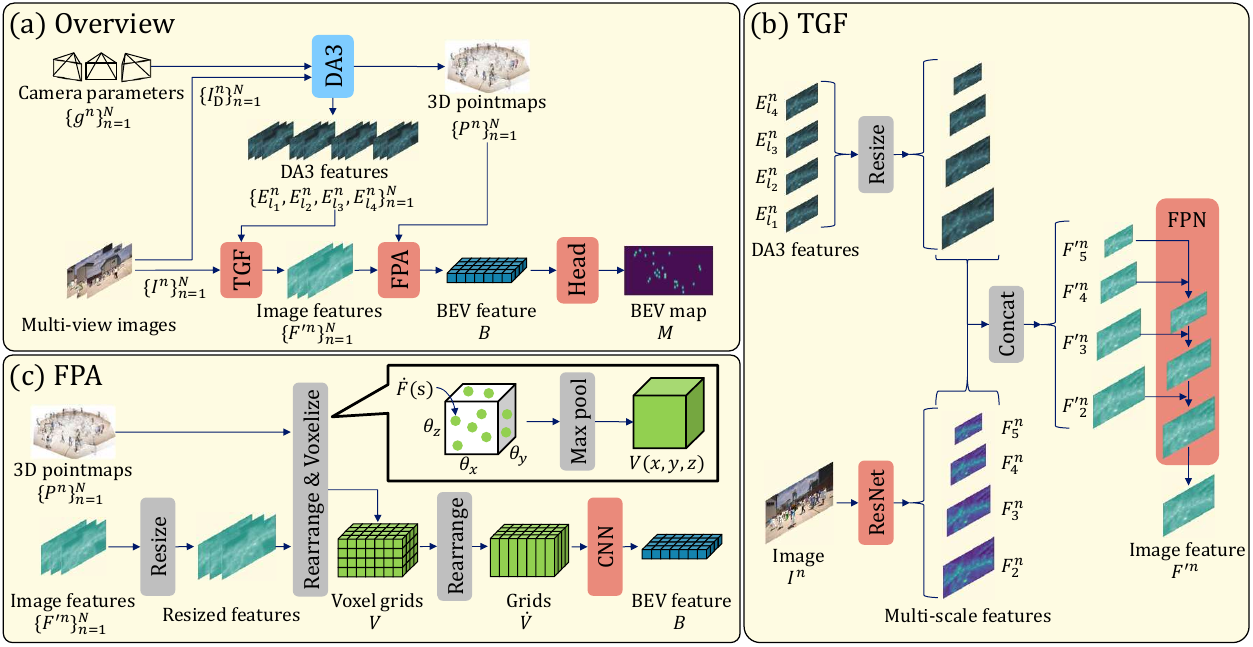}
    \caption{
    (a) Overview of MV2GF. It consists of (b) TGF and (c) FPA. We set the red blocks to be trainable, and the gray blocks do not have trainable parameters.
    }
    \vspace{-2.0mm}
    \label{fig:2}
\end{figure}

\subsection{Visual Geometric Foundation Models}
\label{sub:foundation}

Recently, visual geometric foundation models~\cite{wang2024dust3r,leroy2024grounding,cabon2025must3r,yang2025fast3r,wang2025continuous,tang2025mv,zhang2025flare,wang2025vggt,wang2025pi,depthanything3,keetha2025mapanything}, also known as feed-forward 3D reconstruction transformers, have exhibited strong generalization abilities across various multi-view 3D tasks, such as multi-view depth estimation, 3D reconstruction, and camera pose estimation.
Through training on large-scale multi-view datasets, they learn to capture accurate visual geometry across views for diverse camera configurations by applying transformers~\cite{vaswani2017attention} to multi-view images or videos.
DUSt3R~\cite{wang2024dust3r} and MASt3R~\cite{leroy2024grounding} are pioneers in visual geometric foundation models. 
They predict 3D pointmaps between two views by applying a transformer to a pair of images and compute various 3D attributes from these pointmaps.
While effective for two views, they require a time-consuming and unstable post-processing to consider more than two views.
Subsequent studies have focused on overcoming this limitation~\cite{cabon2025must3r,yang2025fast3r,wang2025continuous,tang2025mv,zhang2025flare,wang2025vggt,wang2025pi,depthanything3,keetha2025mapanything}.
Among them, VGGT~\cite{wang2025vggt} has achieved better results across various 3D tasks than previous methods by applying a transformer to multiple views in a single forward pass and leveraging multi-task learning.
After that, MapAnything~\cite{keetha2025mapanything}, Pi3X~\cite{wang2025pi}, and Depth Anything 3 (DA3)~\cite{depthanything3} have been extended from VGGT to predict 3D attributes at a real-world metric scale and inject camera parameters.
Our MV2GF leverages DA3 because of such abilities and its high generalization performance for diverse camera configurations.

\section{Proposed Method}

This section introduces MV2GF, which leverages a visual geometric foundation model to improve the generalization performance of the detection model to camera configurations not included in MVPD training data.
Figure~\ref{fig:2}~(a) shows an overview of MV2GF that contains TGF and FPA.
We leverage Depth Anything 3 (DA3)~\cite{depthanything3} as a visual geometric foundation model.
From multi-view images and camera parameters, DA3 extracts general-purpose geometric features that capture visual geometry across views and predicts 3D pointmaps for individual views.
MV2GF leverages these geometric features and 3D pointmaps in TGF and FPA, respectively. 
TGF aims to fuse internally extracted task-specific features with general-purpose geometric features extracted by DA3 and to generate informative high-resolution image features.
By doing so, MV2GF uses the visual geometry captured by DA3 for MVPD.
FPA aims to project image features extracted by TGF into a 3D world space using 3D pointmaps predicted by DA3 and to generate a BEV feature by aggregating features across views.
By using these pointmaps, MV2GF effectively avoids shadow-like distortions.
Finally, a detection head predicts a BEV map from the BEV feature.
Hereafter, we explain DA3, TGF, FPA, and the training and inference of MV2GF in that order.

\subsection{Preliminary}
\label{sub:preliminary}

In this subsection, we explain DA3~\cite{depthanything3}, which our method leverages as a visual geometric foundation model.
The inputs to DA3 are multi-view images $\{ I_\mathrm{D}^{n} \}_{n=1}^{N}$ and camera parameters $\{ g^n \}_{n=1}^{N}$ (intrinsics and extrinsics) of $N$ views.
$I_\mathrm{D}^{n} \in \mathbb{R}^{3 \times H_\mathrm{D} \times W_\mathrm{D}}$ is the $n$-th view image, where $H_\mathrm{D}$ and $W_\mathrm{D}$ are the height and width of the input image.
$g^{n} \in \mathbb{R}^9$ is the $n$-th view camera parameters, representing the translation vector $t^{n} \in \mathbb{R}^{3}$, the rotation quaternion $q^{n} \in \mathbb{R}^{4}$, and the field of view $f^{n} \in \mathbb{R}^{2}$.
First, DA3 patchifies $I_\mathrm{D}^{n}$ into a set of tokens $T_{0}^{n} \in \mathbb{R}^{C_\mathrm{D} \times (H_\mathrm{D} / 14 \cdot W_\mathrm{D} / 14)}$ and embeds $g^{n}$ into a token $c_{0}^{n} \in \mathbb{R}^{C_\mathrm{D} \times 1}$, where $C_\mathrm{D}$ is the channel size of DA3.
Then, it captures visual geometry across views via a transformer with $L$ blocks consisting of cross-view and within-view attention, as
\begin{align}
    \centering
    & \{ c_{l}^{n}, T_{l}^{n} \}_{n=1}^{N} = \mathrm{TransformerBlock}_{l} (\{ c_{l-1}^{n}, T_{l-1}^{n} \}_{n=1}^{N}),
\label{formula:transformer}
\end{align}
where $c_{l}^{n} \in \mathbb{R}^{C_\mathrm{D} \times 1}$ and $T_{l}^{n} \in \mathbb{R}^{C_\mathrm{D} \times (H_\mathrm{D} / 14 \cdot W_\mathrm{D} / 14)}$ are tokens from the $l$-th transformer block $\mathrm{TransformerBlock}_{l} (\cdot)$.
For simplicity, we omit the positional encoding in the transformer.
DA3 predicts a 3D pointmap $P^{n} \in \mathbb{R}^{3 \times H_\mathrm{D} \times W_\mathrm{D}}$ for each view using tokens from four blocks $\{ l_1,l_2,l_3,l_4 \}$ and camera parameters, as
\begin{align}
    \centering
    & P^{n} = \mathrm{Prediction} ( \{ T_{l_{1}}^{n}, T_{l_{2}}^{n}, T_{l_{3}}^{n}, T_{l_{4}}^{n}, g^{n} \} ),
\label{formula:pred}
\end{align}
where $ \mathrm{Prediction} (\cdot)$ indicates the prediction head of DA3.
For 2D image pixel coordinates $(u, v)$, $P^{n} (u, v) \in \mathbb{R}^{3}$ represents predicted 3D world coordinates of the pixel $I_\mathrm{D}^{n} (u, v) \in \mathbb{R}^{3}$.
Let $\{ E_{l_{1}}^{n}, E_{l_{2}}^{n}, E_{l_{3}}^{n}, E_{l_{4}}^{n} \}_{n=1}^{N}$ be rearranged multi-view image features, where $E_{l}^{n} \in \mathbb{R}^{C_\mathrm{D} \times H_\mathrm{D} / 14 \times W_\mathrm{D} / 14 }$ is rearranged from $T_{l}^{n}$.
MV2GF uses these $\{ E_{l_{1}}^{n}, E_{l_{2}}^{n}, E_{l_{3}}^{n}, E_{l_{4}}^{n} \}_{n=1}^{N}$ in TGF and the pointmaps $\{ P^{n} \}_{n=1}^{N}$ in FPA.

\subsection{Task-specific and Geometric Information Fusion}
\label{sub:tgf}

Figure~\ref{fig:2}~(b) shows our Task-specific and Geometric information Fusion (TGF).
TGF generates informative high-resolution image features for individual views by fusing task-specific features extracted by a trainable ResNet~\cite{he2016deep} and general-purpose geometric features extracted by the frozen DA3.
The former features highlight foreground pedestrians in each view, and the latter features help capture visual geometry across views for diverse camera configurations.

The inputs to TGF are DA3 features $\{ E_{l_{1}}^{n}, E_{l_{2}}^{n}, E_{l_{3}}^{n}, E_{l_{4}}^{n} \}_{n=1}^{N}$ and multi-view images $\{ I^{n} \}_{n=1}^{N}$.
The input resolution of $I^{n} \in \mathbb{R}^{3 \times H \times W}$ may be different from that of $I_\mathrm{D}^{n}$ due to the pretraining setting of DA3.
First, TGF extracts multi-scale features $\{ F_{2}^{n}, F_{3}^{n}, F_{4}^{n}, F_{5}^{n}\}$ for each view using a trainable ResNet, where $F_{k}^{n} \in \mathbb{R}^{C_{k} \times H/2^{k} \times W/2^{k}}$ and $C_{k}$ are the $n$-th view feature and the channel size of the $k$-th stage in ResNet, respectively.
By making ResNet trainable for MVPD, these features contain task-specific information focused on pedestrians at various scales.
Then, TGF resizes $ E_{l_{1}}^{n}$, $E_{l_{2}}^{n}$, $E_{l_{3}}^{n}$, and $E_{l_{4}}^{n}$ to the same resolution as $ F_{2}^{n}$, $F_{3}^{n}$, $F_{4}^{n}$, and $F_{5}^{n}$ using a bilinear interpolation, respectively, and concatenates the same resolution features along the channel direction.
Let these concatenated features be $\{ F_{2}^{'n}, F_{3}^{'n}, F_{4}^{'n}, F_{5}^{'n}\}$, where $F_{k}^{'n} \in \mathbb{R}^{(C_{k} + C_\mathrm{D})\times H/2^{k} \times W/2^{k}}$.
Finally, by applying FPN~\cite{lin2017feature} to these features, TGF gradually fuses task-specific information with general-purpose geometric information from low-resolution to high-resolution, resulting in informative high-resolution image feature $F^{'n} \in \mathbb{R}^{C \times H/4 \times W/4}$.
Here, $C$ is the channel size in FPN.
By performing these processes for all views, TGF outputs $\{ F^{'n} \}_{n=1}^{N}$.
These features contain not only task-specific information but also general-purpose geometric information extracted by DA3, helping capture visual geometry across views for diverse camera configurations, even those not seen in MVPD training data.

\subsection{Feature Pointmap Aggregation}
\label{sub:fpa}

Figure~\ref{fig:2}~(c) shows our Feature Pointmap Aggregation (FPA).
FPA projects image features extracted by TGF into a 3D world space based on 3D pointmaps predicted by DA3 and aggregates them into a single BEV feature.
Using these pointmaps, each pixel in the TGF features is projected to an appropriate 3D location, thereby avoiding shadow-like distortions.
This makes the detection model more robust to camera configurations not included in the training data than employing a perspective transformation or its derivatives used in existing methods.

The inputs to FPA are 3D pointmaps $\{ P^{n} \}_{n=1}^{N}$ from DA3 and image features $\{ F^{'n} \}_{n=1}^{N}$ from TGF.
First, FPA resizes each of $\{ F^{'n} \}_{n=1}^{N}$ to the same resolution as $P^{n}$ using a bilinear interpolation and rearranges the $N$ pointmaps and resized features into $\dot{P} \in \mathbb{R}^{3 \times (N \cdot H_\mathrm{D} \cdot W_\mathrm{D})}$ and $\dot{F} \in \mathbb{R}^{C \times (N \cdot H_\mathrm{D} \cdot W_\mathrm{D})}$.
After this, we obtain the 3D world location for each $C$-channel feature vector $\dot{F} (i) \in \mathbb{R}^{C}$ by $\dot{P} (i) \in \mathbb{R}^{3}$.
Next, FPA creates $C$-channel voxel grids $V \in \mathbb{R}^{C \times X/4 \times Y/4 \times Z}$ and initializes each voxel with a $C$-channel zero vector.
Here, $X$ and $Y$ are the height and width of a BEV map defined by the datasets, and $Z$ is the assumed vertical dimension.
We set the height and width of $V$ to one-fourth those of the original BEV map for computational efficiency.
Then, for each voxel in $V$, FPA performs max pooling to feature vectors whose 3D locations lie within that voxel and uses the pooling result as a feature vector of the corresponding voxel.
Let the voxel size of $V$ be $\theta_{x} \times \theta_{y} \times \theta_{z}$.
This process is formulated, as
\begin{align}
    \centering
    &S (x, y, z) = \left\{ s \in \{1, \dots, N \cdot H_\mathrm{D} \cdot W_\mathrm{D}\} \;\middle|\; \begin{pmatrix} \theta_{x} x \\ \theta_{y} y \\ \theta_{z} z \end{pmatrix} \le \dot{P}(s) < \begin{pmatrix} \theta_{x} (x+1) \\ \theta_{y} (y+1) \\ \theta_{z} (z+1) \end{pmatrix} \right\}, \\
    &V (x, y, z) = \underset{s \in S (x, y, z)}{\mathrm{max}} \dot{F} (s) ,
\label{formula:pool}
\end{align}
where $(x, y, z)$ indicates the voxel coordinates in $V$.
If $S (x, y, z) = \varnothing$, we keep $V (x, y, z)$ as the zero vector.
Finally, we compress $Z$ dimension in $V$ by rearranging it to $\dot{V} \in \mathbb{R}^{(C \cdot Z) \times X/4 \times Y/4}$ and applying a CNN, resulting in a BEV feature $B \in \mathbb{R}^{C \times X/4 \times Y/4}$.
This series of processes projects the most relevant and informative features from multiple views into each 3D voxel and aggregates information across views into a BEV feature.
FPA projects each feature vector to an appropriate 3D location, avoiding shadow-like distortions and improving generalization to camera configurations not included in the training data.

\subsection{Training and Inference}
\label{sub:train}

The detection head, followed by the sigmoid function, predicts a BEV pedestrian occupancy map $M \in \mathbb{R}^{1 \times X/4 \times Y/4}$ from a BEV feature $B$.
Because $M$ has a lower resolution than the ground truth BEV map, the detection head also regresses the offset map $O \in \mathbb{R}^{2 \times X/4 \times Y/4}$ to recover the discretization error, as in \cite{hou2021multiview}.

During training, we compute Focal loss~\cite{lin2017focal} for $M$ and L1 loss~\cite{zhou2019objects} for $O$, following \cite{hou2021multiview}.
Let $\mathcal{L}_\mathrm{det}$ and $\mathcal{L}_\mathrm{off}$ be losses for $M$ and $O$, respectively.
They are defined using the down-sampled and Gaussian-smoothed ground truth BEV map $\bar{M} \in \mathbb{R}^{1 \times X/4 \times Y/4}$ and the ground truth offset map $\bar{O}\in \mathbb{R}^{2 \times X/4 \times Y/4}$, as
\begin{align}
    \centering
    &\mathcal{L}_\mathrm{det} = \mathrm{FocalLoss} (M, \bar{M}), \ \mathcal{L}_\mathrm{off} = \mathrm{L1Loss} (O, \bar{O}),
\label{formula:det_loss}
\end{align}
where $\mathrm{FocalLoss}(\cdot,\cdot)$ and $\mathrm{L1Loss}(\cdot, \cdot)$ denote Focal loss and L1 loss, respectively.
In addition, as in \cite{hou2020multiview}, we predict occupancy maps of pedestrian heads and feet for each view from TGF image feature $F^{'n}$ and compute Focal loss for these predictions as an auxiliary loss.
Let $M^{n}_{\mathrm{heads}} \in \mathbb{R}^{1 \times H/4 \times W/4}$ and $M^{n}_{\mathrm{feet}} \in \mathbb{R}^{1 \times H/4 \times W/4}$ be predicted occupancy maps of pedestrian heads and feet for $n$-th view. 
We compute a loss $\mathcal{L}^{n}_\mathrm{view}$ for them, as
\begin{align}
    \centering
    &\mathcal{L}^{n}_\mathrm{view} = \mathrm{FocalLoss} (M^{n}_{\mathrm{heads}}, \bar{M}^{n}_{\mathrm{heads}}) + \mathrm{FocalLoss} (M^{n}_{\mathrm{feet}}, \bar{M}^{n}_{\mathrm{feet}}),
\label{formula:aux_loss}
\end{align}
where $ \bar{M}^{n}_{\mathrm{heads}} \in \mathbb{R}^{1 \times H/4 \times W/4}$ and $ \bar{M}^{n}_{\mathrm{feet}} \in \mathbb{R}^{1 \times H/4 \times W/4}$ are the down-sampled ground truth heads and feet map for $n$-th view smoothed with a Gaussian kernel, respectively.
We optimize MV2GF using the following overall loss $\mathcal{L}$, as
\begin{align}
    \centering
    &\mathcal{L} = \mathcal{L}_\mathrm{det} + \mathcal{L}_\mathrm{off} + \frac{1}{N} \Sigma_{n=1}^{N} \mathcal{L}^{n}_\mathrm{view}.
\label{formula:overall_loss}
\end{align}

During inference, we extract pedestrians over the threshold from a predicted BEV occupancy map $M$ and convert their locations to those in a $X \times Y$ BEV map using a predicted offset map $O$, in the same way as \cite{hou2021multiview}.

\section{Experiments}

\subsection{Datasets}
\label{sub:data}

In this subsection, we explain the three MVPD datasets used in our experiments.
Note that the following datasets are not used for pretraining of DA3.

\vskip 0.25\baselineskip
\noindent
\textbf{Wildtrack}~\cite{chavdarova2018wildtrack}.
This is a real-world dataset comprising $400$ multi-view frames at $1080 \times 1920$ resolution captured by $7$ cameras.
Wildtrack covers a $12 \, \mathrm{m} \times 36 \, \mathrm{m}$ region quantized into $480 \times 1440$ grids using square grid cells of $2.5 \, \mathrm{cm} \times 2.5 \, \mathrm{cm}$.
Each frame includes $20$ pedestrians in the region on average.
The first $360$ frames are used for training, and the last $40$ frames are used for testing.

\vskip 0.25\baselineskip
\noindent
\textbf{MVPerception}~\cite{yang2024end}.
This is a synthetic dataset comprising $800$ multi-view frames captured by $6$ cameras and closely follows the style of Wildtrack.
This dataset covers a $16 \, \mathrm{m} \times 25 \, \mathrm{m}$ region quantized into $640 \times 1000$ grids.
Each frame includes $80$ pedestrians in the region on average, making this dataset more crowded than the other two datasets.
The first $720$ frames are used for training, and the remaining $80$ frames are used for testing.

\vskip 0.25\baselineskip
\noindent
\textbf{GMVD}~\cite{vora2023bringing}.
This is a large-scale synthetic dataset.
While the other two datasets consist of a single scene and camera configuration, GMVD comprises $7$ scenes and $13$ camera configurations.
In addition, this dataset contains diverse environmental conditions, such as time and weather.
Therefore, GMVD is more challenging than the other two datasets.
A total of $5997$ multi-view frames at $1080 \times 1920$ resolution are included in this dataset.
Each scene covers a different-sized region, and each region is quantized into grids using the same grid cell size as Wildtrack.
Each frame includes $20$-$40$ pedestrians on average.
In our experiments, we split GMVD in two ways.
First, we split GMVD into $6$ scenes with $11$ camera configurations for training and the remaining $1$ scene with $2$ camera configurations for testing.
We refer to this split as GMVD-D and use it to evaluate detection performance when camera configurations during testing differ from those during training.
Second, we split GMVD into the first $80\%$ frames of each scene and camera configuration for training and the remaining $20\%$ frames for testing.
We refer to this split as GMVD-S and use it to evaluate detection performance when camera configurations during testing are the same as those during training.

\begin{table}[t]
\centering
\scalebox{0.78}{
\begin{tabular}{l|cccc|cccc|cccc} \hline
\multirow{2}{*}{Method} & \multicolumn{4}{c|}{GMVD-D} & \multicolumn{4}{c|}{MVPerception} & \multicolumn{4}{c}{Wildtrack} \\
& MODA & MODP & \ Prec.\  & \ Rec.\ \  & MODA & MODP & \ Prec.\  & \ Rec.\ \  & MODA & MODP & \ Prec.\  & \ Rec.\ \  \\ \hline
MVDet~\cite{hou2020multiview}  & 69.3 & 76.8 & 94.9 & 73.2 & 65.5 & 71.1 & 90.1 & 73.6 & 60.7 & 72.5 & 82.6 & 76.9 \\ 
SHOT~\cite{song2021stacked}    & 71.5 & 77.9 & 93.7 & 76.7 & 69.4 & 73.1 & 91.1 & 76.8 & 70.8 & 73.1 & 82.3 & 90.2 \\
MVDeTr~\cite{hou2021multiview} & 73.8 & \underline{81.7} & 96.1 & 76.9 & 71.3 & 78.0 & 92.8 & 77.3 & 72.4 & 78.1 & 85.6 & 87.1 \\
3DROM~\cite{qiu20223d}        & 74.5 & 77.8 & 93.7 & 80.0 & 74.1 & 75.6 & 91.1 & 82.1 & 76.2 & 75.7 & 85.2 & 92.1 \\
BoosterSHOT~\cite{hwang2024booster} & 74.9 & 80.2 & \underline{96.2} & 77.8 & 74.5 & 76.4 & 87.4 & 87.1 & 78.6 & 79.0 & 87.4 & 91.8 \\
OmniOcc~\cite{aung2024mvpocc}  & 75.1 & 76.9 & 92.3 & 82.0 & 76.6 & 74.0 & 90.3 & 85.7 & 82.0 & 77.9 & 90.5 & 91.7 \\
MVFP~\cite{aung2024enhancing}  & 75.7 & 78.2 & 94.3 & 80.5 & 79.9 & 74.8 & 91.9 & \underline{87.6} & 85.2 & 78.3 & \underline{92.2} & 93.1 \\
MSMVD~\cite{yamane2025msmvd}  & \underline{80.2} & 81.3 & 95.7 & \underline{83.9} & \underline{81.8} & \underline{79.4} & \underline{94.2} & 87.2 & \underline{85.7} & \underline{79.3} & \underline{92.2} & \textbf{93.6} \\ \hline
Ours & \textbf{84.8} & \textbf{83.1} & \textbf{96.8} & \textbf{87.7} & \textbf{86.5} & \textbf{80.6} & \textbf{98.5} & \textbf{87.8} & \textbf{87.9} & \textbf{79.6} & \textbf{94.7} & \underline{93.2} \\ \hline
\end{tabular}
}
\caption{Comparison with previous methods under settings where camera configurations during testing are different from those during training. All models are trained on GMVD-D training split.}
\vspace{-2.0mm}
\label{table:diff}
\end{table}

\subsection{Implementation Details and Evaluation Metrics}
\label{sub:imple}

Unless otherwise mentioned, we used ResNet18~\cite{he2016deep} in TGF and pretrained DA3 Giant~\cite{depthanything3} as a visual geometric foundation model.
ResNet18 was pretrained on ImageNet~\cite{deng2009imagenet}.
DA3 Giant consists of $40$ transformer blocks (\ie, $L$ = 40) and uses tokens from $20$th, $28$th, $34$th, and $40$th blocks for prediction (\ie, $\{ l_{1}, l_{2}, l_{3}, l_{4} \} = \{ 20, 28, 34, 40 \}$).
The channel size $C_\mathrm{D}$ in DA3 was $1536$.
Since the grid cell size of the ground truth BEV map is $2.5 \, \mathrm{cm} \times 2.5 \, \mathrm{cm}$ for all datasets, we set $\theta_{x}$ and $\theta_{y}$ in FPA to $10 \, \mathrm{cm}$.
We set $\theta_{z}$ to $50 \, \mathrm{cm}$ and $Z$ of voxel grids $V$ to $4$.
This means using features up to $2.0 \, \mathrm{m}$ in the vertical direction, covering the height of most pedestrians.
We set the channel size $C$ in FPN to $256$ and the detection threshold to $0.4$.
We used a convolutional layer for the detection head.

We resized input images to $720 \times 1280$ for TGF and to $280 \times 504$ for DA3.
We optimized the model using the Adam optimizer~\cite{loshchilov2018decoupled}.
We set the batch size to $1$ and accumulated gradients over $16$ batches.
We initialized the learning rate to $1.0 \times 10^{-3}$ and decayed it to $1.0 \times 10^{-6}$ following a cosine schedule.
We trained the model for $10$ epochs on GMVD-D and GMVD-S, and for $50$ epochs on Wildtrack and MVPerception.
More detailed implementations are provided in Supp.~A.

Following previous studies~\cite{hou2020multiview,vora2023bringing}, we used four common metrics proposed by Chavdarova~\etal~\cite{chavdarova2018wildtrack} and Kasturi~\etal~\cite{kasturi2008framework}: multiple object detection accuracy (MODA), multiple object detection precision (MODP), precision (Prec.), and recall (Rec.). 
A detected pedestrian was classified as a true positive if its distance from the ground truth was within $0.5$ meters.
MODA was used as the primary metric because it considers both false positives and false negatives.

\subsection{Comparison with Previous Methods}
\label{sub:comparison}

To verify the effectiveness of MV2GF, we compared its generalization ability to camera configurations not included in MVPD training data with that of previous state-of-the-art methods.
Especially, we employed MVDet~\cite{hou2020multiview}, SHOT~\cite{song2021stacked}, MVDeTr~\cite{hou2021multiview}, 3DROM~\cite{qiu20223d}, BoosterSHOT~\cite{hwang2024booster}, OmniOcc~\cite{aung2024mvpocc}, MVFP~\cite{aung2024enhancing}, and MSMVD~\cite{yamane2025msmvd}.
For the first five methods, we added max pooling after the projection to handle any number of views by modifying their official implementations.
For the other methods, we re-implemented them from scratch based on the original papers.
These previous methods rely on a perspective transformation or its derivatives to project image features and learn to capture visual geometry across views solely from camera configurations in MVPD training data.
Table~\ref{table:diff} shows the comparison results when models were trained on GMVD-D training split and tested on GMVD-D testing split, MVPerception, and Wildtrack.
MV2GF outperformed previous methods on all metrics for GMVD-D and MVPerception.
It also achieved the highest MODA, MODP, and precision, and the second-highest recall for Wildtrack.
In particular, it exceeded the previous highest MODA by $4.6$ points for GMVD-D, $4.7$ points for MVPerception, and $2.2$ points for Wildtrack.
These results show that MV2GF generalizes better to camera configurations not included in MVPD training data than previous methods and demonstrate the effectiveness of leveraging a visual geometric foundation model for MVPD.
We visualized the detection results in Supp.~E.

\begin{table}[t]
\centering
\scalebox{0.78}{
\begin{tabular}{l|cccc|cccc|cccc} \hline
\multirow{2}{*}{Method} & \multicolumn{4}{c|}{GMVD-S} & \multicolumn{4}{c|}{MVPerception} & \multicolumn{4}{c}{Wildtrack} \\
& MODA & MODP & \ Prec.\  & \ Rec.\ \  & MODA & MODP & \ Prec.\  & \ Rec.\ \  & MODA & MODP & \ Prec.\  & \ Rec.\ \  \\ \hline
MVDet~\cite{hou2020multiview}       & 82.8 & 80.4 & 97.1 & 85.3 & 90.0 & 81.2 & 94.3 & 95.8 & 88.2 & 75.7 & 94.7 & 93.6 \\ 
SHOT~\cite{song2021stacked}         & 84.5 & 79.6 & 97.5 & 86.7 & 92.3 & 85.5 & 95.2 & \underline{97.1} & 90.2 & 76.5 & 96.1 & 94.0 \\
MVDeTr~\cite{hou2021multiview}      & 86.9 & 84.6 & \underline{97.4} & 89.3 & 93.0 & 90.2 & 96.5 & 96.6 & 91.5 & 82.1 & \underline{97.4} & 94.0 \\
3DROM~\cite{qiu20223d}             & 88.1 & 84.7 & 97.0 & 90.5 & 94.1 & 85.5 & 97.4 & 96.7 & 93.5 & 75.9 & 97.2 & 96.2 \\
BoosterSHOT~\cite{hwang2024booster} & 88.7 & 84.9 & 97.0 & 91.5 & 94.5 & 88.5 & 98.3 & 96.1 & 92.8 & \textbf{84.9} & \textbf{97.5} & 95.3 \\
OmniOcc~\cite{aung2024mvpocc}       & 89.8 & 82.7 & 97.2 & 92.4 & 94.2 & 86.4 & 97.4 & 96.8 & 93.5 & 81.5 & 94.9 & 97.8 \\
MVFP~\cite{aung2024enhancing}       & 88.6 & 81.0 & \underline{97.4} & 91.0 & 95.0 & 85.7 & 98.4 & 96.5 & 94.1 & 78.8 & 96.4 & 97.7 \\
MSMVD~\cite{yamane2025msmvd}        & \underline{91.1} & \textbf{86.6} & \textbf{98.0} & \underline{93.0} & \underline{96.4} & \textbf{91.1} & \textbf{99.7} & 96.8 & \underline{94.6} & \underline{83.3} & 95.9 & \underline{98.8} \\ \hline
Ours & \textbf{91.3} & \underline{86.1}& 97.2 & \textbf{93.9} & \textbf{96.5} & \underline{90.5} & \underline{98.6} & \textbf{97.8} & \textbf{94.7} & 82.9 & 95.8 & \textbf{98.9} \\ \hline
\end{tabular}
}
\caption{Comparison with previous methods under settings where camera configurations during testing are identical to those during training. All models are trained on the training split of the same dataset used for testing.}
\vspace{-2.0mm}
\label{table:same}
\end{table}

While our primary objective is to improve the generalization ability of the detection model, detection performance when camera configurations during testing are identical to those during training is also important.
Table~\ref{table:same} shows the comparison results when models were trained on GMVD-S training split, MVPerception training split, or Wildtrack training split and tested on the corresponding testing split.
For all datasets, MV2GF achieved superior MODA and recall to all previous methods and was comparable to MSMVD.
Particularly on MODA, it outperformed MSMVD by $0.2$, $0.1$, and $0.1$ points for GMVD-S, MVPerception, and Wildtrack, respectively.
These results demonstrate that MV2GF is effective even when camera configurations during testing are identical to those during training.
On the other hand, MV2GF underperformed MSMVD on MODP and precision.
This is because MSMVD uses multi-scale BEV features with multi-scale image features, enabling precise pedestrian localization.
In Supp.~C, we investigated the effect of this multi-scale BEV design.

\begin{table}[t]
\centering
\begin{tabular}{cc}
\scalebox{0.90}{
\begin{tabular}{cc|cccc} \hline
TGF & FPA & MODA & MODP & \ Prec.\  & \ Rec.\ \ \\ \hline
           &            & 73.4 & 78.5 & 93.2 & 79.1 \\ 
\checkmark &            & 77.8 & 79.5 & 93.8 & 83.3 \\
           & \checkmark & 81.3 & 82.6 & 95.5 & 85.4 \\
\rowcolor{gray!20}
\checkmark & \checkmark & \textbf{84.8} & \textbf{83.1} & \textbf{96.8} & \textbf{87.7} \\ \hline
\end{tabular}
}&
\scalebox{0.90}{
\begin{tabular}{cc|cccc} \hline
ResNet & DA3 & MODA & MODP & \ Prec.\  & \ Rec.\ \ \\ \hline
\checkmark &            & 81.3 & 82.6 & 95.5 & 85.4 \\
           & \checkmark & 77.5 & 77.9 & 95.1 & 81.7 \\
\rowcolor{gray!20}
\checkmark & \checkmark & \textbf{84.8} & \textbf{83.1} & \textbf{96.8} & \textbf{87.7} \\ \hline
\end{tabular}
} \\
(a)&(b) \\
\scalebox{0.90}{
\begin{tabular}{l|cccc} \hline
& MODA & MODP & \ Prec.\  & \ Rec.\ \ \\ \hline
Only 20th  & 83.6 & 82.7 & 96.5 & 86.7 \\ 
Only 28th  & 83.6 & 83.0 & 96.4 & 86.8 \\ 
Only 34th  & 83.9 & 83.0 & 96.5 & 87.1 \\ 
Only 40th  & 84.0 & \textbf{83.1} & 96.0 & 87.6 \\
\rowcolor{gray!20}
All & \textbf{84.8} & \textbf{83.1} & \textbf{96.8} & \textbf{87.7} \\ \hline
\end{tabular}
}&
\scalebox{0.90}{
\begin{tabular}{l|cccc} \hline
$\theta_{z} \, (\mathrm{cm})$ & MODA & MODP & \ Prec.\  & \ Rec.\ \ \\ \hline
12.5 & 81.6 & 81.7 & 96.3 & 84.8 \\ 
25   & 84.5 & 82.8 & 96.6 & 87.5 \\
\rowcolor{gray!20}
50   & \textbf{84.8} & \textbf{83.1} & \textbf{96.8} & \textbf{87.7} \\
\rowcolor{white}
100  & 84.3 & \textbf{83.1} & 96.3 & 87.6 \\
200  & 83.7 & 82.9 & 96.1 & 87.3 \\ \hline
\end{tabular}
} \\
(c)&(d) \\
\scalebox{0.90}{
\begin{tabular}{l|cccc} \hline
& MODA & MODP & \ Prec.\  & \ Rec.\ \ \\ \hline
DINOv2 & 82.9 & 82.2 & 95.6 & 86.9 \\ 
DINOv3 & 83.2 & 82.5 & 96.6 & 86.2 \\ 
\rowcolor{gray!20}
DA3    & \textbf{84.8} & \textbf{83.1} & \textbf{96.8} & \textbf{87.7} \\ \hline
\end{tabular}
}&
\scalebox{0.90}{
\begin{tabular}{l|cccc} \hline
& MODA & MODP & \ Prec.\  & \ Rec.\ \ \\ \hline
MapAnything & 83.0 & 82.3 & 96.5 & 86.2 \\
Pi3X        & 82.3 & 82.6 & 96.4 & 85.5 \\
\rowcolor{gray!20}
DA3         & \textbf{84.8} & \textbf{83.1} & \textbf{96.8} & \textbf{87.7} \\ \hline
\end{tabular}
} \\
(e)&(f) \\
\end{tabular}
\caption{(a) Effect of TGF and FPA. (b) Effect of using features from ResNet and DA3 in TGF. (c) Effect of using features from multiple transformer blocks in DA3. (d) Effect of $\theta_{z}$ in FPA. (e) Effect of using features from a visual geometric foundation model. (f) Effect of the visual geometric foundation model choice. Default settings in MV2GF are marked in \colorbox{gray!20}{gray}. All models are trained on GMVD-D training split and tested on its testing split.}
\label{tab:ablation}
\end{table}

\subsection{Ablation Studies}
\label{sub:alation}

In this subsection, we investigated the effects of each component in MV2GF from various viewpoints.
We conducted more detailed investigations in Supp.~B.
Unless otherwise mentioned, all models were trained on GMVD-D training split and tested on GMVD-D testing split.

\vskip 0.25\baselineskip
\noindent
\textbf{Effect of TGF and FPA}.
To investigate the effects of TGF and FPA on detection performance, we incrementally added each of them to a baseline model.
As a baseline model, we used MVDet~\cite{hou2020multiview} trained with our loss and employed ResNet18~\cite{he2016deep} and FPN~\cite{lin2017feature} for its image feature extraction.
Table~\ref{tab:ablation}~(a) shows the effects of TGF and FPA.
Adding TGF improved detection performance on all metrics, and the improvement was $4.4$ points on MODA.
The baseline with TGF outperformed all previous methods in Tab.~\ref{table:diff} on MODA, other than MSMVD.
This result demonstrates the effectiveness of using geometric features from DA3 via TGF to capture visual geometry across views.
Adding FPA also improved detection performance, and the improvement was greater than that by TGF.
In addition, the baseline with FPA outperformed all previous methods in Tab.~\ref{table:diff} on MODA.
This result highlights the effectiveness of projecting image features via FPA while avoiding shadow-like distortions.
The combined use of TGF and FPA further improved detection performance, and the improvement from the baseline was $11.4$ points on MODA.
This result indicates that both TGF and FPA are important for better generalization to camera configurations not included in MVPD training data.

\vskip 0.25\baselineskip
\noindent
\textbf{Effect of using features from ResNet and DA3 in TGF}.
In TGF, we utilized both task-specific ResNet features and general-purpose geometric DA3 features.
To verify the advantage of this, we compared cases where both features were used with only ResNet features or DA3 features were used, as shown in Tab.~\ref{tab:ablation}~(b).
Using both features outperformed using either on all metrics. 
In particular, using both features outperformed using only ResNet features and only DA3 features by $3.5$ points and $7.3$ points on MODA, respectively.
These results demonstrate that both task-specific ResNet features and general-purpose geometric DA3 features are important for better detection performance in camera configurations not included in the training data.

\vskip 0.25\baselineskip
\noindent
\textbf{Effect of using features from multiple transformer blocks in DA3}.
In TGF, we used geometric DA3 features from four transformer blocks ($20$th, $28$th, $34$th, and $40$th blocks) that were also used for predicting 3D pointmaps in DA3.
To verify the advantage of this, we compared cases where features from all four blocks were used with features only from one block were used, as shown in Tab.~\ref{tab:ablation}~(c).
Using features from four blocks yielded better detection performance across all metrics than using features from a single block.
This result demonstrates the effectiveness of capturing visual geometry across views using the features from multiple transformer blocks in DA3.

\vskip 0.25\baselineskip
\noindent
\textbf{Effect of $\theta_{z}$ in FPA}.
We compared different $\theta_{z}$ in FPA, as shown in Tab.~\ref{tab:ablation}~(d).
Note that we set $\theta_{z}$ to $50 \, \mathrm{cm}$ by default, and all models used features up to $2.0 \, \mathrm{m}$ in the vertical direction.
From $200 \, \mathrm{cm}$ to $50 \, \mathrm{cm}$, smaller $\theta_{z}$ achieved better detection performance.
This is because a too large $\theta_{z}$ compresses features too much in the vertical direction, preventing the use of the entire pedestrian's body information in predicting a BEV map.
On the other hand, from $50 \, \mathrm{cm}$ to $12.5 \, \mathrm{cm}$, employing smaller $\theta_{z}$ degraded detection performance.
This is because a too small $\theta_{z}$ leaves many grids in $V$ as zero vectors, making $V$ too redundant.
These results show that our default $\theta_{z} = 50 \, \mathrm{cm}$ is the best choice.

\vskip 0.25\baselineskip
\noindent
\textbf{Effect of using features from a visual geometric foundation model}.
To verify that the performance improvement by using DA3 features stems from its characteristic tailored to capturing visual geometry across views rather than merely from a large amount of pretrained data, we implemented cases where DA3 features in TGF were replaced with features of DINOv2~\cite{oquab2023dinov2} or DINOv3~\cite{simeoni2025dinov3}, as shown in Tab.~\ref{tab:ablation}~(e).
When using DINOv2 or DINOv3 features, superior detection performance was achieved compared to using only ResNet features (see Tab.~\ref{tab:ablation}~(b)), but it was inferior to that of using DA3 features.
This result demonstrates the advantage of using DA3 features that focus on capturing visual geometry across views.

\vskip 0.25\baselineskip
\noindent
\textbf{Effect of the visual geometric foundation model choice}.
In the experiments, MV2GF employed DA3 as a visual geometric foundation model because it can predict 3D attributes at a real-world metric scale, inject calibrated camera parameters, and process more than two views in a single forward pass.
MapAnything~\cite{keetha2025mapanything} and Pi3X~\cite{wang2025pi} also have such abilities, while DA3 predicts 3D attributes more accurately than them.
We investigated the effect of the visual geometric foundation model choice by replacing geometric features in TGF and 3D pointmaps in FPA with those of MapAnything or Pi3X, as shown in Tab.~\ref{tab:ablation}~(f).
Employing DA3 achieved better detection performance than employing the others.
This result indicates that employing a stronger visual geometric foundation model yields better detection performance.
On the other hand, even when MapAnything or Pi3X was employed, our method outperformed all previous methods in Tab.~\ref{table:diff}.
This demonstrates that our method generalizes better than previous methods, regardless of the visual geometric foundation model choice.

\begin{table}[t]
\centering
\scalebox{0.90}{
\begin{tabular}{c|cccc|cccc} \hline
& \multicolumn{4}{c|}{MSMVD} & \multicolumn{4}{c}{Ours} \\
\ \ $N$ \ \ & MODA & MODP & \ Prec.\  & \ Rec.\ \  & MODA & MODP & \ Prec.\ & \ Rec.\ \ \\ \hline
\ \ 6 \ \ & 82.7 & 81.1 & 95.2 & 87.1 & 85.9 & 82.0 & 96.8 & 88.9  \\
\ \ 5 \ \ & 79.9 & 79.0 & 93.0 & 86.3 & 84.1 & 80.5 & 96.8 & 87.0  \\
\ \ 4 \ \ & 70.5 & 78.7 & 88.6 & 80.9 & 80.0 & 80.3 & 95.5 & 84.0  \\
\ \ 3 \ \ & 61.4 & 72.6 & 85.0 & 74.5 & 73.8 & 78.0 & 93.7 & 79.1  \\
\ \ 2 \ \ & 34.0 & 63.0 & 76.0 & 49.8 & 57.6 & 73.3 & 86.3 & 67.6  \\ \hline 
\end{tabular}
}
\caption{Effect of the number of input views $N$ for MSMVD~\cite{yamane2025msmvd} and our MV2GF. Both models are trained on GMVD-D training split and tested on data comprising $6$ views in GMVD-D testing split.}
\vspace{-2.0mm}
\label{table:views}
\end{table}

\vskip 0.25\baselineskip
\noindent
\textbf{Effect of the number of views}.
We investigated the effect of the number of input views $N$ on detection performance.
Table~\ref{table:views} shows the detection performance of MSMVD~\cite{yamane2025msmvd} and our MV2GF for different $N$ on data comprising $6$ views in GMVD-D testing split.
Reducing $N$ negatively affected detection performance across all metrics for both MSMVD and MV2GF.
Meanwhile, MV2GF exhibited less performance degradation than MSMVD and achieved superior detection performance across all $N$.
This result demonstrates that MV2GF is more robust than MSMVD with respect to the number of input views.

\section{Limitations}
\label{sec:limit}

Although leveraging a visual geometric foundation model yields high generalization performance for MV2GF to camera configurations not included in MVPD training data, it has a negative impact on inference speed.
To investigate this impact, we measured the inference speed of our MV2GF and previous methods using an Nvidia 80GB A100 GPU.
Table~\ref{table:speed} shows the inference speed of each model, including data loading and preprocessing.
These inference speeds were measured on data comprising $6$ views in GMVD-D testing split.
MV2GF was slightly slower than previous methods.
In particular, the gap in inference speed between MV2GF and previous methods was from $1.2$ FPS to $1.9$ FPS. 
While the inference speed required for practical application has not been explicitly discussed in the field of MVPD, most MVPD datasets comprise multi-view videos at a frame rate of $2.0$ FPS~\cite{chavdarova2018wildtrack,hou2020multiview,vora2023bringing,yang2024end}.
Therefore, the inference speed of MV2GF is sufficient to process these datasets.
In future work, we will explore an effective way to accelerate the inference of MV2GF by employing a model compression technique (\eg, knowledge distillation).

\begin{table}[t]
\centering
\scalebox{0.90}{
\begin{tabular}{l|c} \hline
& \ FPS \ \\ \hline
MVDet~\cite{hou2020multiview}       & 5.2  \\
SHOT~\cite{song2021stacked}         & 4.8  \\
MVDeTr~\cite{hou2021multiview}      & 4.6  \\
3DROM~\cite{qiu20223d}              & 4.6  \\
BoosterSHOT~\cite{hwang2024booster} & 4.7  \\
OmniOcc~\cite{aung2024mvpocc}       & 4.6  \\
MVFP~\cite{aung2024enhancing}       & 4.6  \\
MSMVD~\cite{yamane2025msmvd}        & 4.5  \\
Ours                                & 3.3  \\ \hline 
\end{tabular}
}
\caption{Comparison of inference speed. These inference speeds are measured on data comprising $6$ views in GMVD-D testing split using an Nvidia 80GB A100 GPU.}
\vspace{-2.0mm}
\label{table:speed}
\end{table}

\section{Discussion and Conclusion}

We propose a new multi-view pedestrian detection (MVPD) method, named MV2GF.
It leverages a visual geometric foundation model to improve the generalization ability of the detection model to camera configurations not included in MVPD training data.
In particular, MV2GF leverages multi-view features and 3D pointmaps from this foundation model.
The former contains general-purpose geometric information and helps capture visual geometry across views for diverse camera configurations, even those not seen in MVPD training data.
The latter is used to project each pixel in image features to an appropriate location in a 3D world space, thereby avoiding shadow-like distortions that occur in previous methods.
Extensive experiments demonstrated the advantage of leveraging a visual geometric foundation model for MVPD, and MV2GF achieved significantly superior generalization performance compared to previous methods.
We hope that our work will broaden the scope of MVPD applications.

When MV2GF was trained on data with different camera configurations from those in the testing data (see Tab.~\ref{table:diff}), it generalized better than previous methods.
However, its detection performance still underperformed previous methods trained on data with the same camera configurations as the testing data (see Tab.~\ref{table:same}).
Optimizing detection performance for specific camera configurations is important for some applications.
In future work, we will explore a way to further improve detection performance in such situations while maintaining generalization ability to camera configurations not included in MVPD training data.



%
%
\bibliographystyle{splncs04}
\bibliography{main}

\clearpage
\appendix


{
\centering
\Large
\textbf{MV2GF: Multi-view Pedestrian Detection \\ with a Visual Geometric Foundation Model}\\
\vspace{0.5em}Supplementary Material \\
\vspace{0.5em}
}

\setcounter{page}{1}
\setcounter{table}{5}
\setcounter{figure}{2}

\begin{table}[ht]
\centering
\begin{tabular}{cc}
\scalebox{0.90}{
\begin{tabular}{l|cccc} \hline
& MODA & MODP & \ Prec.\  & \ Rec.\ \ \\ \hline
w/o offset & 84.4 & 79.0 & 96.5 & 87.5 \\ 
\rowcolor{gray!20}
w/ offset  & \textbf{84.8} & \textbf{83.1} & \textbf{96.8} & \textbf{87.7} \\ \hline
\end{tabular}
}&
\scalebox{0.90}{
\begin{tabular}{l|cccc} \hline
& MODA & MODP & \ Prec.\  & \ Rec.\ \ \\ \hline
w/o $\mathcal{L}^{n}_\mathrm{view}$ & 83.8 & 82.0 & 95.8 & 87.6 \\
\rowcolor{gray!20}
w/ $\mathcal{L}^{n}_\mathrm{view}$  & \textbf{84.8} & \textbf{83.1} & \textbf{96.8} & \textbf{87.7} \\ \hline
\end{tabular}
} \\
(a)&(b) \\
\scalebox{0.90}{
\begin{tabular}{l|cccc} \hline
& MODA & MODP & \ Prec.\  & \ Rec.\ \ \\ \hline
\rowcolor{gray!20}
ResNet18  & 84.8 & 83.1 & 96.8 & 87.7 \\
\rowcolor{white}
ResNet50  & 85.0 & 83.7 & \textbf{97.4} & 87.3 \\
ResNet101 & \textbf{85.5} & \textbf{84.1} & \textbf{97.4} & \textbf{87.9} \\ \hline
\end{tabular}
}&
\scalebox{0.90}{
\begin{tabular}{l|cccc} \hline
Pooling & MODA & MODP & \ Prec.\  & \ Rec.\ \ \\ \hline
Mean & 84.5 & 81.8 & \textbf{96.8} & 87.3 \\
\rowcolor{gray!20}
Max  & \textbf{84.8} & \textbf{83.1} & \textbf{96.8} & \textbf{87.7} \\ \hline
\end{tabular}
} \\
(c)&(d) \\
\scalebox{0.90}{
\begin{tabular}{l|cccc} \hline
& MODA & MODP & \ Prec.\  & \ Rec.\ \ \\ \hline
DINOv2  & 78.7 & 76.5 & 96.0 & 82.1 \\
DINOv3  & 79.4 & 77.8 & 96.5 & 82.4 \\
\rowcolor{gray!20}
ResNet & \textbf{84.8} & \textbf{83.1} & \textbf{96.8} & \textbf{87.7} \\ \hline
\end{tabular}
}&
\scalebox{0.90}{
\begin{tabular}{l|cccc} \hline
Pooling & MODA & MODP & \ Prec.\  & \ Rec.\ \ \\ \hline
w/ loss weighting & 84.2 & 83.1 & 96.6 & 87.0 \\
\rowcolor{gray!20}
w/o loss weighting  & \textbf{84.8} & 83.1 & \textbf{96.8} & \textbf{87.7} \\ \hline
\end{tabular}
} \\
(e)&(f) \\
\end{tabular}
\caption{(a) Effect of the offset prediction. (b) Effect of auxiliary loss $\mathcal{L}^{n}_\mathrm{view}$. (c) Effect of ResNet size in TGF. (d) Effect of the pooling choice in FPA. (e) Effect of using features from a trainable ResNet. (f) Effect of loss weighting. Default settings in MV2GF are marked in \colorbox{gray!20}{gray}. All models are trained on GMVD-D training split and tested on its testing split.}
\vspace{-4.0mm}
\label{tab:supp_ablation}
\end{table}

\section{Implementation Details}
\label{sec:supp_imple}

For a CNN in FPA, we used three dilated convolutional layers~\cite{Yu2016}, as in \cite{hou2020multiview}.
For the prediction of each view pedestrian heads occupancy map $M_\mathrm{heads}^{n}$ and feet occupancy map $M_\mathrm{feet}^{n}$ during training, we added an additional convolutional layer and applied it to TGF image feature $F^{'n}$.
For the Adam optimizer~\cite{loshchilov2018decoupled}, we did not use the weight decay and set the optimizer momentum to $\beta_1 = 0.9$ and $\beta_2 = 0.99$.
We set the diameter of the Gaussian kernel to $10$ pixels for the down-sampled ground truth BEV map $\bar{M}$, heads maps $\{ \bar{M}_{\mathrm{heads}}^{n} \}_{n=1}^{N}$, and feet maps $\{ \bar{M}_{\mathrm{feet}}^{n} \}_{n=1}^{N}$.
We conducted all experiments on an Nvidia $80$GB A100 GPU.

In our MV2GF, DA3 has $1.4$B parameters, and the trainable modules have $23.0$M parameters.
For GPU memory, our method uses $27.8$ GB during training on GMVD-D by pre-computing DA3 outputs and $21.5$ GB, including DA3, during inference on $6$ views.

\section{Additional Ablation Studies}
\label{sec:supp_ablation}

\begin{table}[t]
\centering
\scalebox{0.90}{
\begin{tabular}{l|cccc} \hline
& MODA & MODP & \ Prec.\  & \ Rec.\ \ \\ \hline
MSMVD~\cite{yamane2025msmvd}       & 91.1 & 86.6 & 98.0 & 93.0 \\
\rowcolor{gray!20}
Ours w/o MSMVD's multi-scale BEV design & 91.3 & 86.1 & 97.2 & 93.9 \\
\rowcolor{white}
Ours w/ MSMVD's multi-scale BEV design  & \textbf{91.8} & \textbf{87.2} & \textbf{98.1} & \textbf{94.0} \\ \hline
\end{tabular}
}
\caption{Effect of MSMVD's multi-scale BEV design on GMVD-S. All models are trained on GMVD-S training split and tested on GMVD-S testing split. Camera configurations during training are identical to those during testing. Default settings in MV2GF are marked in \colorbox{gray!20}{gray}.}
\vspace{-4.0mm}
\label{table:multiscale}
\end{table}

In this section, we conducted more detailed investigations into the effects of each component in MV2GF from various viewpoints.
All models were trained on GMVD-D training split and tested on GMVD-D testing split.

\vskip 0.5\baselineskip
\noindent
\textbf{Effect of offset prediction}.
Several previous methods~\cite{hou2020multiview,song2021stacked,aung2024enhancing,aung2024mvpocc,zhang2024mahala} do not predict the offset map and ignore the discretization errors caused by down-sampling of the BEV map.
To investigate the effect of using the offset map, we compared MV2GF with and without the offset prediction, as shown in Tab.~\ref{tab:supp_ablation}~(a).
MV2GF with the offset prediction achieved better detection performance on all metrics than without it.
Particularly on MODP, which considers the distance between a predicted pedestrian and the ground truth, the former outperformed the latter by $4.1$ points.
This result demonstrates the necessity of the offset prediction.

\vskip 0.5\baselineskip
\noindent
\textbf{Effect of auxiliary loss $\mathcal{L}^{n}_\mathrm{view}$}.
As described in Sec.~\ref{sub:train}, we trained MV2GF with the per-view auxiliary loss $\mathcal{L}^{n}_\mathrm{view}$.
To investigate this effect, we compared two models trained with and without $\mathcal{L}^{n}_\mathrm{view}$, as shown in Tab.~\ref{tab:supp_ablation}~(b).
The model trained with $\mathcal{L}^{n}_\mathrm{view}$ achieved better detection performance than the model trained without it.
Especially, the former outperformed the latter by $1.0$ points on MODA.
This result demonstrates the advantage of training with the auxiliary loss $\mathcal{L}^{n}_\mathrm{view}$ and indicates that making image features from TGF more discriminative through per-view supervision improves detection performance.

\vskip 0.5\baselineskip
\noindent
\textbf{Effect of ResNet size in TGF}.
In TGF, we extracted task-specific features using ResNet18 for fair comparisons with previous methods.
We applied different sizes of ResNet in TGF and investigated their effects on detection performance, as shown in Tab.~\ref{tab:supp_ablation}~(c).
Employing a larger ResNet achieved better detection performance.
In particular, employing the largest ResNet101 outperformed employing the smallest ResNet18 by $0.7$ points on MODA.
This result indicates that employing a larger ResNet yields more effective task-specific features and improves the generalization ability of the detection model to camera configurations not included in MVPD training data.

\vskip 0.5\baselineskip
\noindent
\textbf{Effect of pooling choice in FPA}.
In FPA, although we applied max pooling to feature vectors within each voxel, we could apply mean pooling instead.
We compared the detection performance with these two pooling methods, as shown in Tab.~\ref{tab:supp_ablation}~(d).
Applying max pooling achieved better detection performance than applying mean pooling.
In particular, the former outperformed the latter by $0.3$, $1.3$, and $0.4$ points on MODA, MODP, and recall, respectively.
Max pooling brings the most relevant and informative features from multiple views for each voxel in $V$ , leading to superior detection performance.

\vskip 0.5\baselineskip
\noindent
\textbf{Effect of using features from ResNet}.
In TGF, we used a trainable ResNet to extract task-specific features.
To verify the advantage of this, we implemented cases where ResNet features in TGF were replaced with features of DINOv2 or DINOv3, as shown in Tab.~\ref{tab:supp_ablation}~(e).
When using DINOv2 or DINOv3 features, the detection performance was slightly superior to using only DA3 features (in Tab.~\ref{tab:ablation}~(b)), but significantly inferior to using ResNet features.
This result demonstrates the advantage of using features extracted by a trainable ResNet that contain task-specific information.

\vskip 0.5\baselineskip
\noindent
\textbf{Effect of loss weighting}.
Following MVDet~\cite{hou2020multiview} and MVDeTr~\cite{hou2021multiview}, we equally weighted $\mathcal{L}_\mathrm{det}$, $\mathcal{L}_\mathrm{off}$, and the auxiliary loss, as described in Sec.~\ref{sub:train}.
The average loss values of $\mathcal{L}_\mathrm{det}$, $\mathcal{L}_\mathrm{off}$, and the auxiliary loss term in Eq.~\ref{formula:overall_loss} during training were $0.35$, $0.18$, and $1.08$ on GMVD-D, respectively.
To investigate the effect of loss weighting, we applied weights to each loss term so that these average values were equal, as shown in Tab.~\ref{tab:supp_ablation}~(f).
This weighting worsened MODA, precision, and recall compared to our default weighting.
This result indicates that our choice is better.

\begin{table}[t]
\centering
\scalebox{0.90}{
\begin{tabular}{l|cccc} \hline
& MODA & MODP & \ Prec.\  & \ Rec.\ \ \\ \hline
MVDet~\cite{hou2020multiview}  & 17.0 & 65.8 & 60.5 & 48.8 \\
SHOT~\cite{song2021stacked}    & 53.6 & 72.0 & 75.2 & 79.8 \\
MVDeTr~\cite{hou2021multiview} & 50.2 & 69.1 & 74.0 & 77.3 \\
3DROM~\cite{qiu20223d}         & 67.5 & 65.6 & \textbf{94.5} & 71.7 \\
MVFP~\cite{aung2024enhancing}  & \underline{82.6} & 76.2 & 89.6 & 93.4 \\
MSMVD~\cite{yamane2025msmvd}   & 82.0 & \underline{78.0} & 88.7 & \underline{94.0} \\
Ours  & \textbf{84.3} & \textbf{78.1} & \underline{90.0} & \textbf{95.0} \\ \hline
\end{tabular}
}
\caption{Comparison with previous methods under the setting where models are trained on MultiviewX and tested on Wildtrack.}
\vspace{-4.0mm}
\label{table:multiviewx}
\end{table}

\section{Effect of MSMVD's multi-scale BEV design}
\label{sec:multiscale}

In Sec.~\ref{sub:comparison}, we explained that the reason why MV2GF underperformed MSMVD on MODP and precision under the setting where camera configurations during testing are identical to those during training is MSMVD's multi-scale BEV design.
To confirm the effect of MSMVD's multi-scale BEV design, we implemented a model that incorporates it into MV2GF under the setting in Tab.~\ref{table:same}.
Table~\ref{table:multiscale} shows the detection performance of that model on GMVD-S.
By incorporating MSMVD's multi-scale BEV design, MV2GF greatly improved MODP and precision, achieving better detection performance than MSMVD on all metrics.
This result demonstrates that MSMVD's multi-scale BEV design is also effective for our method and indicates its importance, particularly for MODP and precision.

\section{MultiviewX to Wildtrack setting}
\label{sec:multiviewx}

In several previous studies~\cite{vora2023bringing,aung2024enhancing}, the generalization performance of detection models to unseen camera configurations during training has been evaluated under the setting where the models were trained on MultiviewX~\cite{hou2020multiview} and tested on Wildtrack~\cite{chavdarova2018wildtrack}.
Since MultiviewX is smaller than GMVD-D, this setting can assess the generalization performance when the training data are less diverse and contain fewer scenes.
We compared our MV2GF with previous methods under this setting, as shown in Tab.~\ref{table:multiviewx}.
For the result of MSMVD, we re-implemented it from scratch based on the original paper~\cite{yamane2025msmvd}.
For the other previous methods, we directly copied their results from the original MVFP paper~\cite{aung2024enhancing}.
MV2GF achieved the highest MODA, MODP, and recall, and the second-highest precision.
This result demonstrates that our method remains effective even when the training data are less diverse and contain fewer scenes.

\section{Visual Comparison of Detection Results}
\label{sec:supp_vis}

\begin{figure}[tb]
    \centering
    \includegraphics[width=100mm]{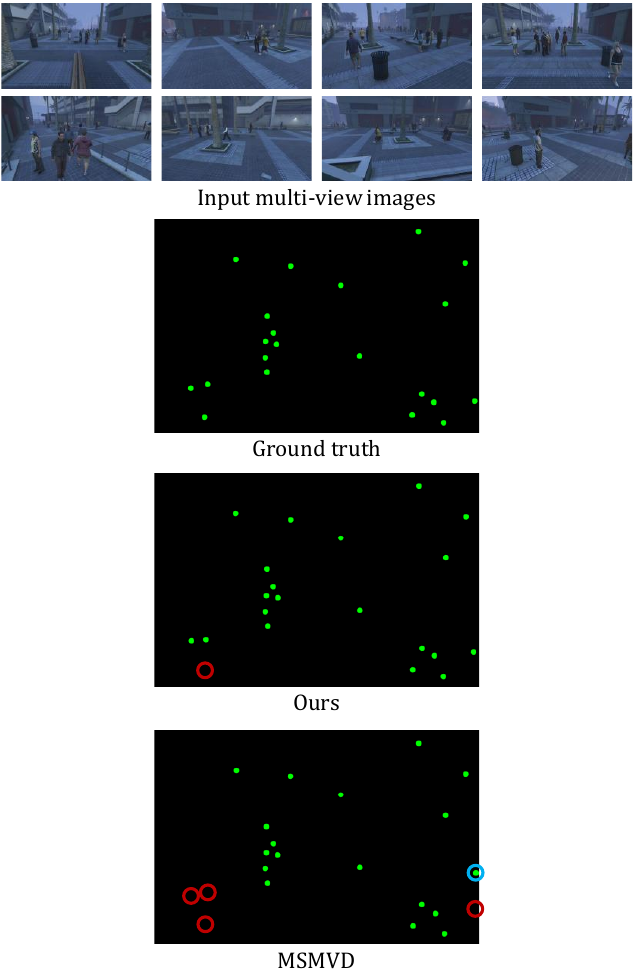}
    \caption{
    Visual comparison of detection results (\ie, predicted BEV maps) between our MV2GF and MSMVD~\cite{yamane2025msmvd} on GMVD-D. The green points represent detected pedestrians, the blue circles represent false positives, and the red circles represent false negatives.
    }
    \vspace{-3.0mm}
    \label{fig:3}
\end{figure}

\begin{figure}[tb]
    \centering
    \includegraphics[width=100mm]{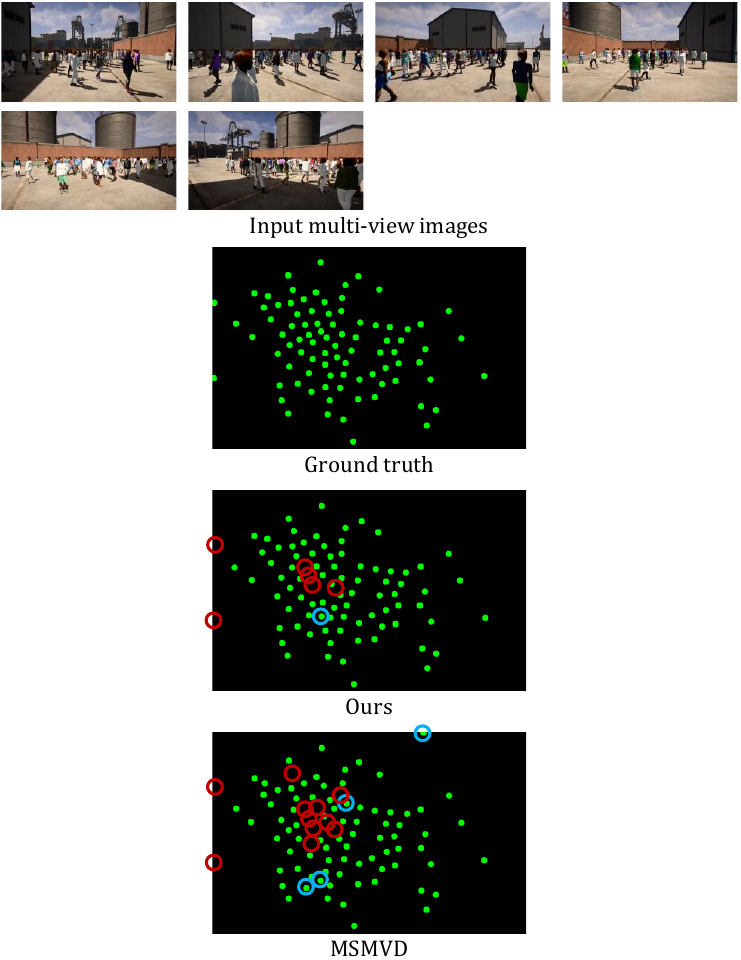}
    \caption{
    Visual comparison of detection results (\ie, predicted BEV maps) between our MV2GF and MSMVD~\cite{yamane2025msmvd} on MVPerception. The green points represent detected pedestrians, the blue circles represent false positives, and the red circles represent false negatives.
    }
    \vspace{-3.0mm}
    \label{fig:4}
\end{figure}

\begin{figure}[tb]
    \centering
    \includegraphics[width=100mm]{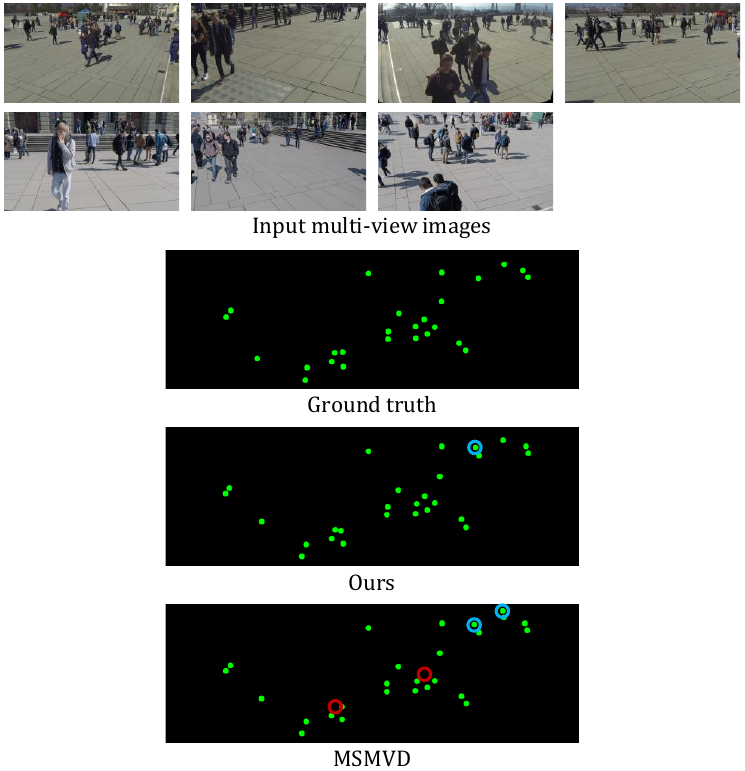}
    \caption{
    Visual comparison of detection results (\ie, predicted BEV maps) between our MV2GF and MSMVD~\cite{yamane2025msmvd} on Wildtrack. The green points represent detected pedestrians, the blue circles represent false positives, and the red circles represent false negatives.
    }
    \vspace{-3.0mm}
    \label{fig:5}
\end{figure}

To visually verify the effectiveness of our MV2GF, we visualized the detection results (\ie, predicted BEV maps) of MV2GF and MSMVD~\cite{yamane2025msmvd}.
Figures~\ref{fig:3}, \ref{fig:4}, and \ref{fig:5} show examples of the detection results in cases where models were trained on GMVD-D training split and were tested on GMVD-D testing split, MVPerception, and Wildtrack, respectively.
In these figures, the green points, the blue circles, and the red circles represent detected pedestrians, false positives, and false negatives, respectively.
MV2GF reduced both false positives and false negatives across all testing datasets.
These results demonstrate that MV2GF generalizes better to camera configurations not included in MVPD training data than MSMVD.

\section{Visualization of 3D pointmaps predicted by DA3}
\label{sec:supp_point}

\begin{figure}[tb]
    \centering
    \includegraphics[width=100mm]{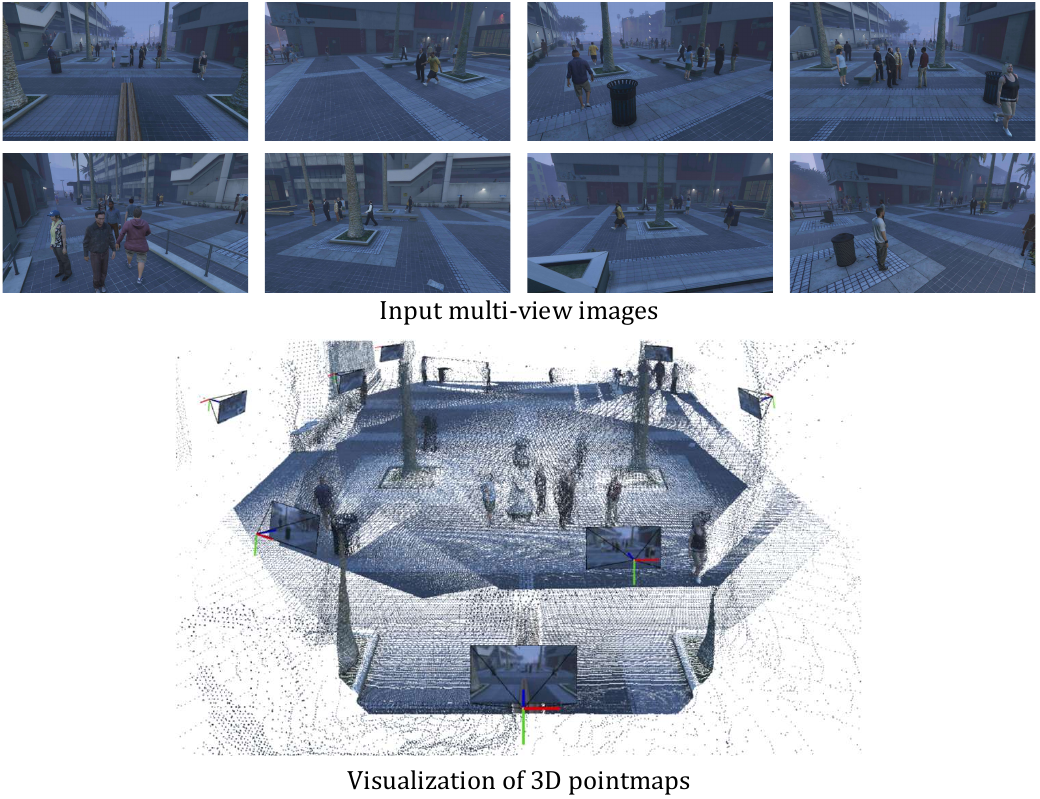}
    \caption{
    Visualization of 3D pointmaps predicted by DA3~\cite{depthanything3} on GMVD.
    }
    \vspace{-3.0mm}
    \label{fig:6}
\end{figure}

\begin{figure}[tb]
    \centering
    \includegraphics[width=100mm]{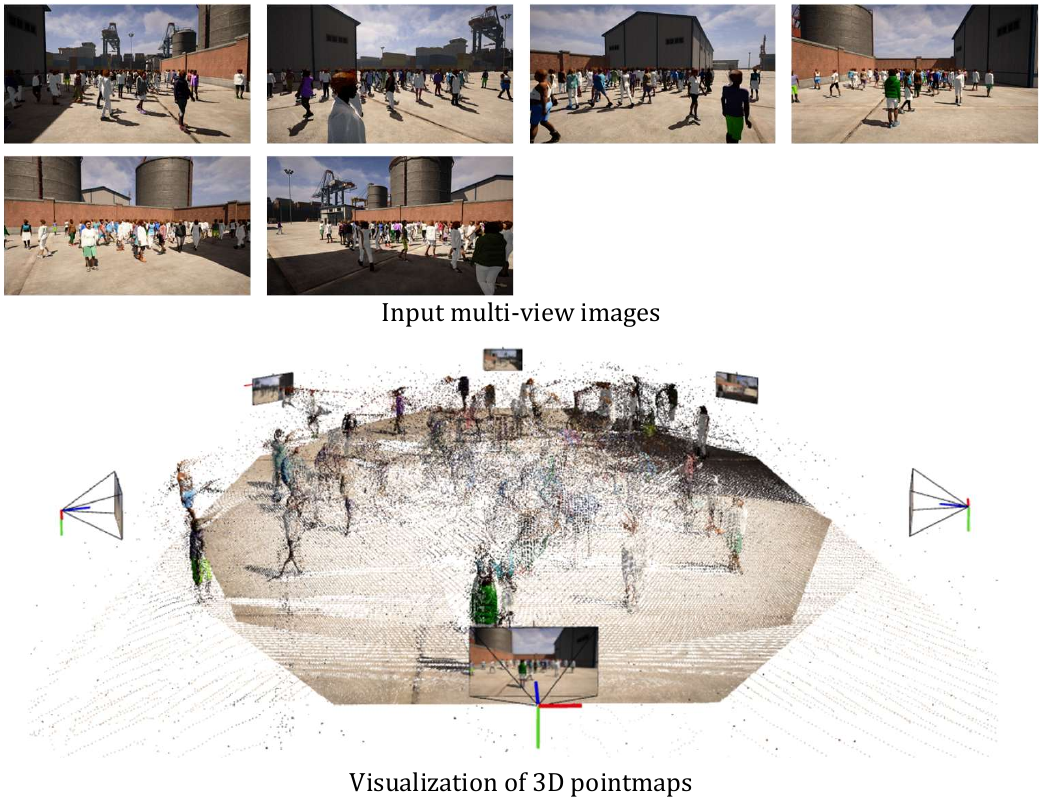}
    \caption{
    Visualization of 3D pointmaps predicted by DA3~\cite{depthanything3} on MVPerception.
    }
    \vspace{-3.0mm}
    \label{fig:7}
\end{figure}

\begin{figure}[tb]
    \centering
    \includegraphics[width=100mm]{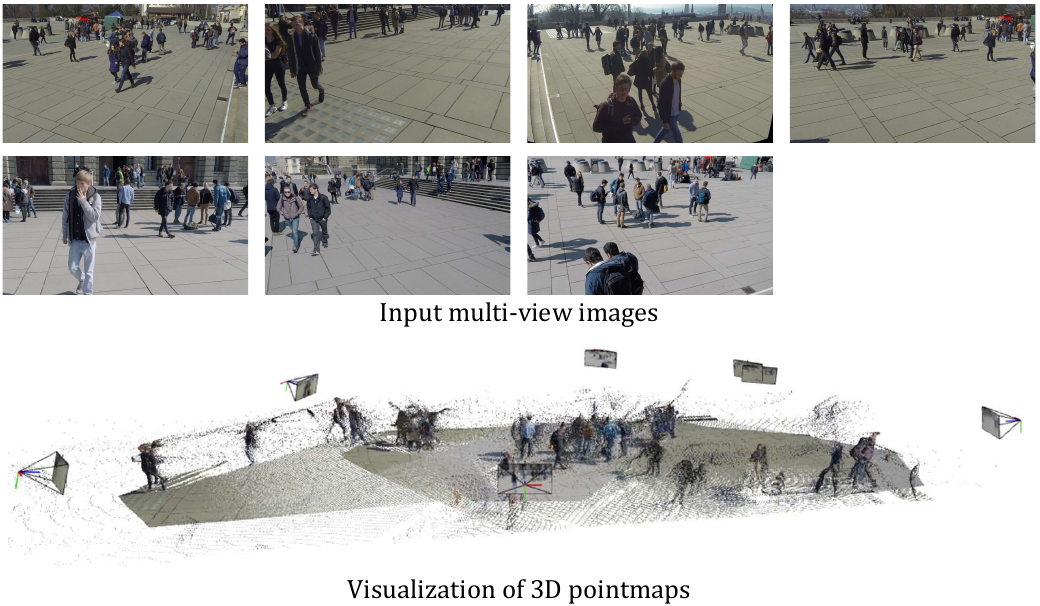}
    \caption{
    Visualization of 3D pointmaps predicted by DA3~\cite{depthanything3} on Wildtrack.
    }
    \vspace{-3.0mm}
    \label{fig:8}
\end{figure}

We visualized 3D pointmaps predicted by DA3.
Figures~\ref{fig:6}, \ref{fig:7}, and \ref{fig:8} show examples of the 3D pointmaps predicted by DA3 on GMVD, MVPerception, and Wildtrack, respectively.
DA3 predicted high-quality 3D pointmaps across all datasets.
These high-quality pointmaps enable MV2GF to project each pixel in 2D image features to an appropriate 3D location, avoiding shadow-like distortions that occur in previous methods.

\section{Visual Comparison of BEV features}
\label{sec:supp_bev}

\begin{figure}[tb]
    \centering
    \includegraphics[width=100mm]{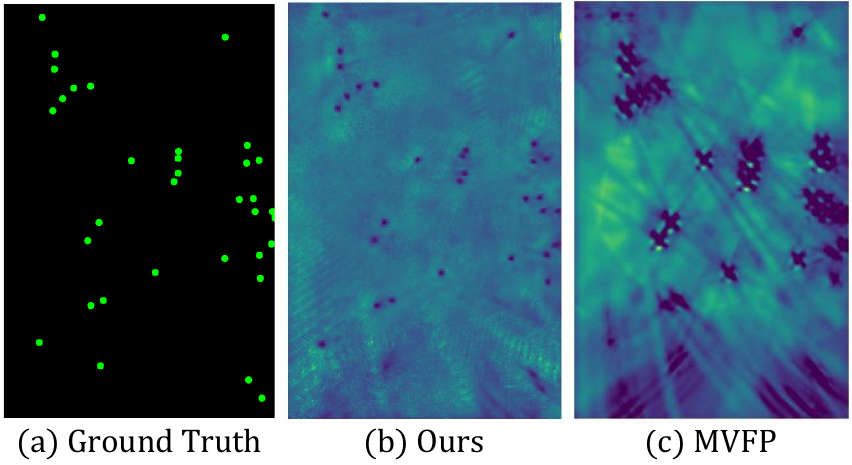}
    \caption{
    Visualization of (a) the ground truth BEV map and BEV features of (b) our MV2GF and (c) MVFP~\cite{aung2024enhancing} on GMVD-D.
    }
    \vspace{-3.0mm}
    \label{fig:9}
\end{figure}

To visually verify that our MV2GF avoids shadow-like distortions, we visualized BEV features of MV2GF and MVFP~\cite{aung2024enhancing}.
Figure~\ref{fig:9} shows (a) the ground truth BEV map and BEV features of (b) our MV2GF and (c) MVFP on GMVD-D.
In the ground truth BEV map, the green points represent pedestrians.
Our MV2GF exhibited fewer distortions around pedestrians than MVFP and precisely represented pedestrian locations.
This result shows that our method avoids shadow-like distortions by using 3D pointmaps predicted by a visual geometric foundation model to project 2D image features into a 3D world space.

\end{document}